\documentclass[a4paper]{jpconf}

\pdfoutput=1

\usepackage{graphicx}
\usepackage{caption}
\usepackage{subcaption}

\usepackage{color}

\usepackage{amsfonts, amsmath, amsthm, amssymb} 

\usepackage{citesort}

\begin{document}
\title{Muon Trigger for Mobile Phones}

\author{
	M~Borisyak$^{1, 2}$,
	M~Usvyatsov$^{2, 3, 4}$,
	M~Mulhearn$^{5}$,
	C~Shimmin$^{6,7}$ and
	A~Ustyuzhanin$^{1, 2}$
}

\address{$^{1}$ National Research University Higher School of Economics, 20 Myasnitskaya st., Moscow 101000, Russia}
\address{$^{2}$ Yandex School of Data Analysis, 11/2, Timura Frunze st., Moscow 119021, Russia}
\address{$^3$~Moscow~Institute~of~Physics~and~Technology, 9 Institutskiy per., Dolgoprudny, Moscow Region, 141700, Russia}
\address{$^4$~Skolkovo~Institute~of~Science~and~Technology, Skolkovo Innovation Center, Building 3, Moscow 143026, Russia}
\address{$^5$~University~of~California,~Davis, 1 One Shields Avenue, Davis, CA 95616, USA}
\address{$^6$~University~of~California,~Irvine, 4129 Frederick Reines Hall, Irvine, CA 92697-4575, USA}
\address{$^7$~Yale~University, 217 Prospect Street, New Haven, CT 06520, USA}

\ead{mborisyak@hse.ru}

\begin{abstract}
The CRAYFIS experiment proposes to use privately owned mobile phones as a ground detector array for Ultra High Energy Cosmic Rays. Upon interacting with Earth's atmosphere, these events produce extensive particle showers which can be detected by cameras on mobile phones. A typical shower contains minimally-ionizing particles such as muons. As these particles interact with CMOS image sensors, they may leave tracks of faintly-activated pixels that are sometimes hard to distinguish from random detector noise. Triggers that rely on the presence of very bright pixels within an image frame are not efficient in this case.

We present a trigger algorithm based on Convolutional Neural Networks which selects images containing such tracks and are evaluated in a lazy manner: the response of each successive layer is computed only if activation of the current layer satisfies a continuation criterion. Usage of neural networks increases the sensitivity considerably comparable with image thresholding, while the lazy evaluation allows for execution of the trigger under the limited computational power of mobile phones.
\end{abstract}

\section{Introduction}
\label{sec:introduction}
The problem of pattern detection over a set of images arises in many applications.
The CRAYFIS experiment is dedicated to observations of Ultra-High-Energy Cosmic Rays (UHECR) by a distributed network of mobile phones provided by volunteers.
In the process of interaction with the Earth's atmosphere, UHECRs produce cascades of particles called Extensive Air Showers (EAS).
Some of the particles reach the ground, affecting areas of up to several hundreds of meters in radius.
These particles can be detected by cameras on mobile phones, and a localized coincidence of particle detection by several phones can be used to observe very rare UHECR events~\cite{whiteson2016searching}.

This approach presents a number of challenges.
In order to observe an EAS, each active smartphone needs to continuously monitor its camera output by scanning megapixel-scale images at rates of 15-60 frames per second.
This generates a vast amount of raw data, which presents problems both
for volunteers\footnote{For example, it may quickly exceed smartphone's storage capacity or introduce a considerable load on networks.} and experimenters if transmitted to data processing servers for later analysis.
However, the recorded data contains almost entirely random camera noise, as signals from cosmic ray interactions are expected to occur in fewer than 1 in 10,000 image frames.
As there would be potentially millions of smartphones operating simultaneously, it is critical to utilize the local processing power available on each device to select only the most interesting data.
Hence, a trigger algorithm is required to filter out background data and identify possible candidates for cosmic rays traces.
It is also important that the camera monitoring is subject to negligible dead time; therefore any trigger must operate with an average evaluation response time on the order of 30ms to track with the raw data rate.

Some constituents of an EAS, such as electrons and gamma-ray photons, leave bright traces in the camera \cite{whiteson2016searching}.
In this case, the simplest trigger strategy is cut on brightness (if there are bright pixels in a fragment then this fragment is considered interesting). This usually enough to provide acceptable background rejection rate in the case of bright traces, and given a target background rejection rate it is possible to automatically determine the threshold value for decision making.
However, this strategy is much less effective against another component of the shower, comprising minimally-ionizing particles such as high-energy muons.
These particles may leave relatively faint signals in the pixels they traverse, possibly at a level comparable to the sensor's intrinsic noise.

Nevertheless, these minimally-ionizing particles traverse the sensor's pixels in distinctively straight lines. If these tracks are long enough in the plane of the sensor, there is a low probability of the same pattern emerging from intrinsic random camera noise. Thus it is still possible to discriminate even these faintly-interacting particles from background.

In this work, we propose a novel approach for fast visual pattern detection, realized here as a trigger designed for fast identification of muon traces on mobile phone's camera.
This method is based on Convolutional Neural Networks and does not contain any specific assumptions for identification of muon traces, hence, in principle, it can be applied to any visual pattern detection problem.



\section{Related Work}
	

Minimally-ionizing particles are characterized by the pattern of activated pixels they leave over a small region of an exposure. Hence the problem of minimally-ionizing particle detection can be transformed to the problem of pattern detection over the image.
Several attempts were performed to solve pattern detection problem in different setups. The solution proposed in works \cite{li2015convolutional} and \cite{ren2015faster} utilizes Convolutional Neural Networks (CNNs).
Certain properties of CNNs such as translation invariance, locality, and correspondingly few weights to be learned, make them particularly well suited to extracting features from images.
The performance of even simple CNNs on image classification tasks have been shown to be very good compared to other methods~\cite{lecun1995convolutional}.
However, the training and evaluation of CNNs requires relatively intense computation to which smartphones are not particularly well suited.
Viola and Jones in \cite{viola2001rapid} introduced the idea of a simple feature-based cascading classifier. They proposed using small binary filters as feature detectors and to increase computational power from cascade to cascade.

Bagherinzhad et al. in \cite{lcnn} enumerated a wide range of methods which have been proposed to address efficient training and inference in deep neural networks. Although these methods are orthogonal to our approach, they may be incorporated with the method described here in order to improve efficiency in related tasks.

\section{CNN trigger}
The key insight of the proposed method is to view a Deep Convolutional Neural Network (CNN) as a chain of triggers, or cascades:
each trigger filters its input stream of data before passing it further down the chain.
The main feature of such chains is that amount of data passing successfully through the chain gradually decreases,
while the complexity of triggers gradually increases, allowing finer selection with each subsequent trigger.
This architecture allows one to effectively control the computational complexity of the overall chain, and usually, to substantially decrease the amount of computational resources required \cite{viola2001rapid}.

Convolutional Neural Networks are particularly well suited for adopting this approach as instead of passing an image itself throughout the chain,
the CNN computes a series of intermediate representations (activations of hidden layers) of the image \cite{lecun1995convolutional}.
Following the same reasoning as in Deeply Supervised Nets (DSN) \cite{lee2015deeply},
one can build a network for which discriminative power of intermediate representations grows along the network, making it possible to treat such CNN as a progressive trigger chain \footnote{However, in contrast to DSN, the growth of discriminative power is a requirement for an effective trigger chain rather than for network regularization.}.

In order to build a trigger chain from a CNN, we propose a method similar to DSN:
each layer of the CNN is extended with a binary trigger based on the image representation obtained by this layer.
In the present work we use logistic regression as model for the trigger, although, any binary classifier with a differentiable model would be sufficient.

\begin{figure}[h]
	\centering
	\begin{subfigure}{0.27\linewidth}
		\includegraphics[width=\linewidth]{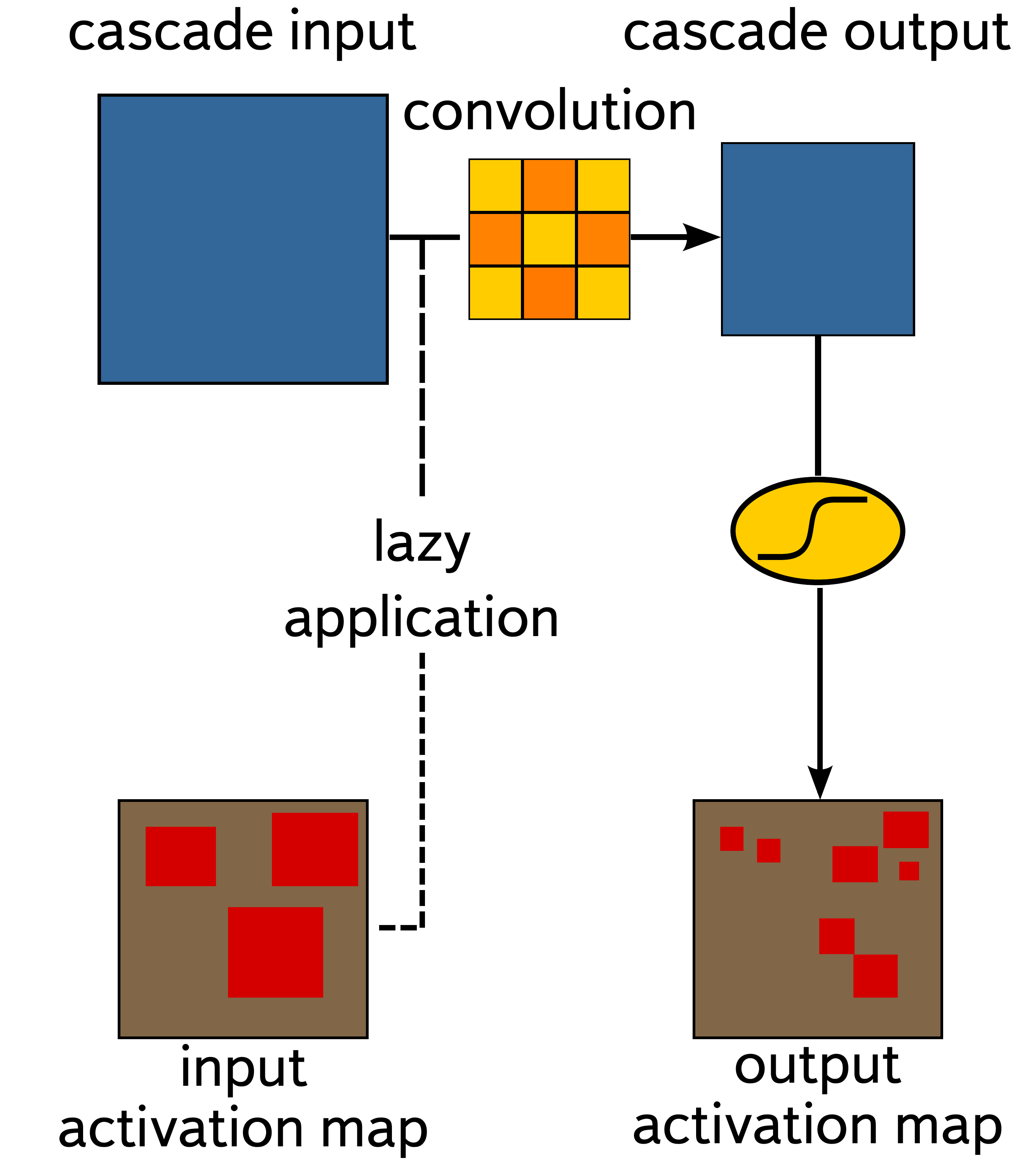}
		\caption{convolutional cascade}
		\label{fig:cascade}
	\end{subfigure}
	~
	\begin{subfigure}{0.66\linewidth}
		\includegraphics[width=\linewidth]{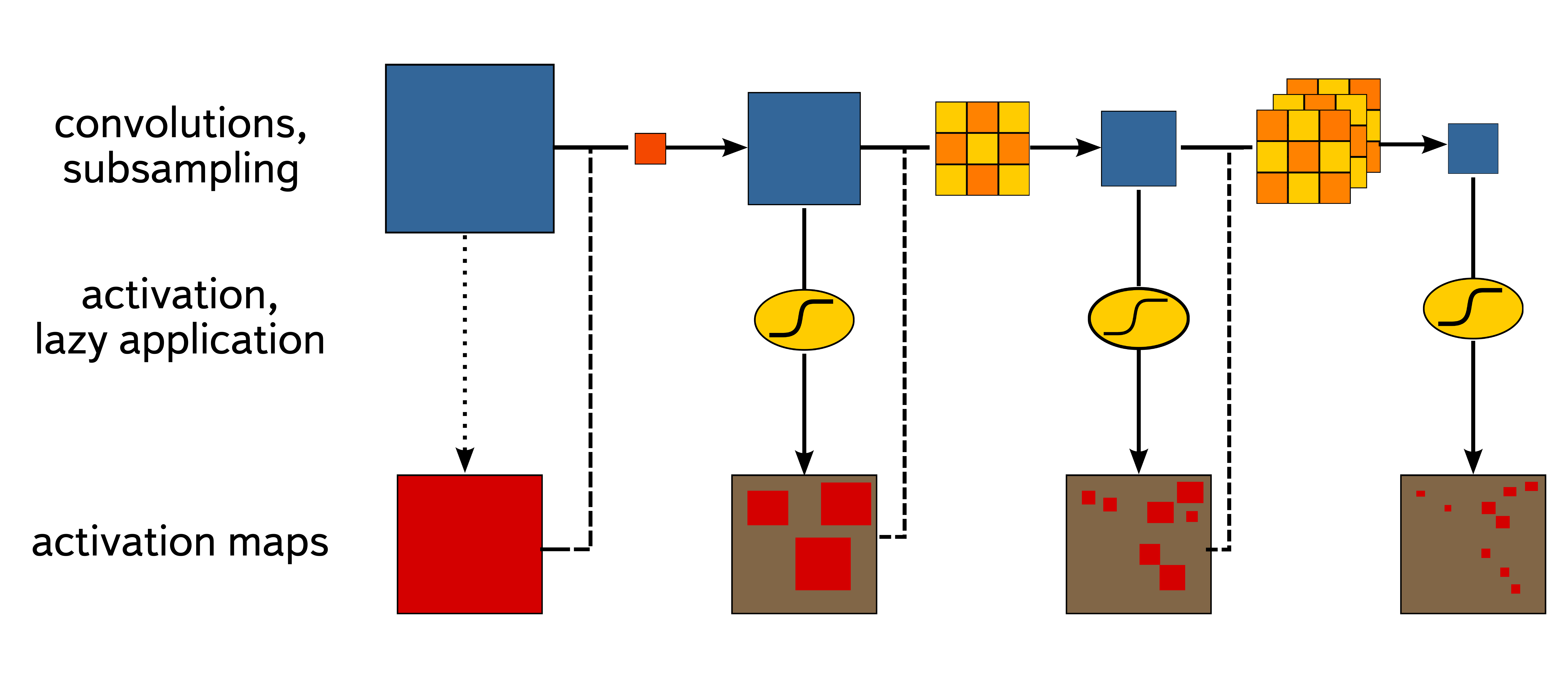}
		\caption{CNN trigger structure}
		\label{fig:trigger}
	\end{subfigure}
	\caption{
		Fig.~\ref{fig:cascade} shows the building block of the CNN trigger, an individual convolutional cascade.
		In contrast to conventional convolutional layers, the convolutional cascade has an additional input, the activation map, which
		indicates regions to which the convolutional operator should be applied (lazy application, denoted by dashed lines).
		The activation map is updated by the associated trigger (represented by the s-shaped node),
		which may be passed on to the subsequent cascade or interpreted as the final output indicating regions of interest.
    		Fig.~\ref{fig:trigger} shows the full structure of CNN trigger as a sequence of convolutional cascades.
	   	Initially the whole image is activated (red areas).
	   	As the image proceeds through the chain, the activated area becomes smaller as each cascade refines the activation map.
    }
    \label{fig:convtrigger}
\end{figure}

The trigger is applied to each region of the output image to determine if that region should proceed further down the trigger chain.
We call these layers with their corresponding triggers a convolutional cascade, by analogy with Viola-Jones cascades \cite{viola2001rapid}.
The output of the trigger at each stage produces what we refer to as an \emph{activation map}, to be passed to the next layer, as illustrated in Fig. \ref{fig:cascade}.

From another perspective, this approach can be seen as an extension of the CNN architecture in which network computations concerning regions of the input image are
halted as soon as a negative outcome for a region becomes evident\footnotemark.
This is effectively accomplished by generalizing the convolution operator to additionally accept as input an activation map indicating the regions to be computed.
In the case where sparse regions of activation are expected, this lazy application can result in much fewer computations being performed.

\footnotetext{Lazy application can be viewed as a variation of attention mechanisms recently proposed
in Deep Learning literature, see e.g. \cite{xu2015show}.}

After each application of the \emph{lazy convolution}, the activation map for the subsequent layer is furnished by the trigger associated with the current layer.
The whole chain is illustrated in Fig.~\ref{fig:trigger}.


Training of the CNN trigger may be problematic for gradient methods, since prediction is no longer a
continuous function of network parameters.
This is because the lazy convolution, described above, is in general non-differentiable.

In order to overcome this limitation, we propose to use a slightly different network architecture during training by substituting a differentiable approximation of the lazy convolution operator.
The basic idea is that instead of skipping the evaluation of unlikely regions, we simply ensure that any region which has low activation on a given cascade will continue to have low activation on all subsequent cascades.
In this scheme, the evaluation is no longer lazy, but since training may be performed on much more powerful hardware, this is not a concern.

\begin{figure}[h]
	\centering
	\includegraphics[width=0.75\linewidth]{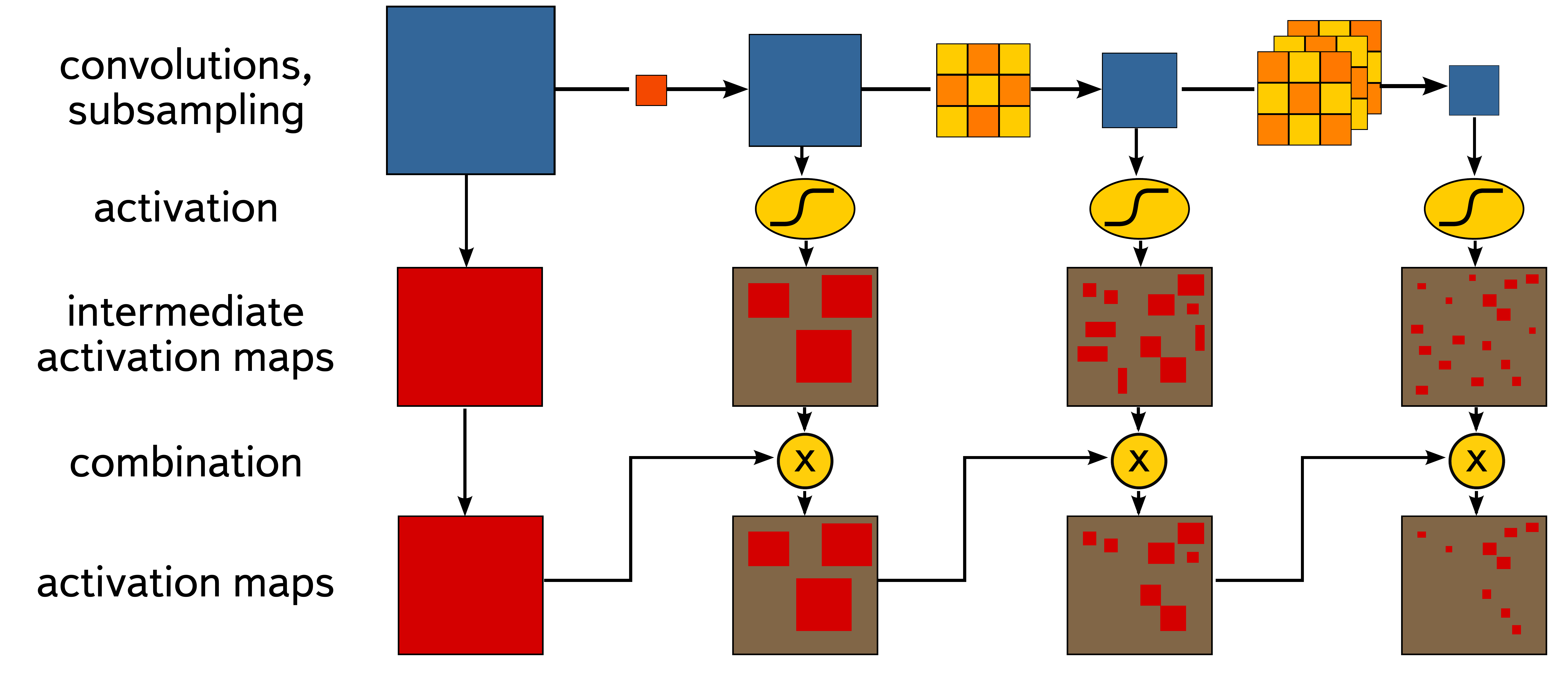}
	\caption{
		Schematic of the CNN trigger used for training.
		To make the network differentiable, lazy application is replaced by
		its approximation, that does not involve any ``laziness''.
		Activation maps are approximated
		as the elementwise product of unconditional trigger response (intermediate activation maps) and the previous activation map.
	}
	\label{fig:training}
\end{figure}

To accomplish this, we first replace the binary activation maps (which result from the trigger classification) with continuous probability estimates.
Secondly, we introduce \emph{intermediate activation maps}, which are computed by the trigger function at each layer.
The intermediate map is multiplied by the previous layer's activation map to produce a refined activation map\footnotemark.
In this way, the activation probability for any region is nonincreasing with each cascade layer.
The process is depicted schematically in Fig.~\ref{fig:training}.

This differentiable version of the lazy application operation for the $i^\mathrm{th}$ cascade is described by the following equations:
\begin{eqnarray}
	I^{i} &=& h^i(I^{i - 1});\\
	\hat{A}_{x, y}^i &=& \sigma^i(I^i_{x, y});\\
	A^i &=& \hat{A}^i \otimes A^{i - 1};\label{eq:activation}\\
	A^0_{x, y} &:=& 1,
\end{eqnarray}
where $I^i$ is the intermediate representation of the input image $I^0$ after successive applications of CNN layers, and $h^i$ represents the transformation associated with the $i^\mathrm{th}$ layer of the CNN (typically this would be convolution, nonlinearity, and pooling).
$\sigma^i$ is the function associated with the trigger (in our case, logistic regression), and its result $\hat{A}^i$ is the intermediate activation map of the $i^\mathrm{th}$ cascade.
Finally, $A^i$ is the differentiable version of the activation map, given by the element-wise product of the intermediate activation with the previous layer's activation.
That elements of the initial activation map $A^0$ are set to 1, and the subscripts $x,y$ denote region position.

Note that the dimensions of the activation map define the granularity of regions-of-interest that may be triggered, and may in general be smaller than the dimensions of the input image.
In this case, the trigger function $\sigma$ should incorporate some downsampling such as max-pooling.

\footnotetext{
	If layers of underlying CNN contains pooling, i.e. change size of the image, pooling should be
	applied to intermediate activation maps as well.
}

We also note that since similar but still technically different networks are used for training and prediction,
special care should be taken while transitioning from probability estimations to binary classification.
In particular, additional adjustment of classifier thresholds may be required.
Nevertheless, in the present work no significant differences in the two networks' behaviors were found (see Fig.~\ref{fig:actmap}).

To train the network we utilize cross-entropy loss.
Since activation maps are also intermediate results, and the activation map $A^n$ of the last cascade is the final result for a network of $n$ cascades, the loss can be written as:
\begin{equation}
	\mathcal{L}^n = -\frac{1}{W \times H} \sum_{x, y} Y_{x, y} \log I^n_{x, y} + \gamma^n (1 - Y_{x, y}) \log (1 - I^n_{x, y})
	\label{eq:cascadeloss}
\end{equation}
where $Y \in \mathbb{R}^{W \times H}$ denotes the ground truth map with width $W$ and height $H$.
The truth map is defined with $Y_{x,y}=1$ if the region at coordinates $(x,y)$ contains a target object, otherwise $Y_{x,y}=0$.
The coefficient $\gamma^n$ is introduced to provide control over the penalty for triggering on background.

If the cross-entropy term \eqref{eq:cascadeloss} is the only component of the loss, the network will have no particular incentive to assign regions a low activation on early cascades, limiting the benefit of the lazy evaluation architecture.
One approach to force network to produce good results on intermediate layers is to directly introduce penalty term $C$ for unnecessary computations:
\begin{equation}
	C = \sum^{n}_{i = 1} c^i \sum_{x, y} \left(1 - Y_{x, y}\right) A^{i - 1}_{x, y} \label{eq:complexity}
\end{equation}
where $c^i$ represents the per-region cost of performing convolution and trigger in the $i^\mathrm{th}$ cascade.

We use a naive estimation of coefficients the $c^i$, assuming, for simplicity, that convolution is performed by
elementwise multiplication of the convolutional kernel with a corresponding image patch.
In this case, for $l$ filters of size $(k, k)$ applied to image with $m$ channels:
\begin{equation}
	c^i = \underbrace{m l k^2}_{\text{multiplications}} + \underbrace{l (m k^2 - 1)}_{\text{summations}} + \underbrace{2l - 1}_{\text{trigger}}
\end{equation}

Combining these terms, the resulting total loss function is given by:
\begin{equation}
	\mathcal{L} = \mathcal{L}^n + \beta C \label{eq:loss1}
\end{equation}
where the parameter $\beta$ is introduced to regulate the trade-off between computational efficiency and classification quality.

Another approach is to apply a DSN technique:
\begin{equation}
	\mathcal{L} = \mathcal{L}^n + \sum^{n - 1}_{i = 1} \alpha^i \mathcal{L}^i \label{eq:loss2}\ .
\end{equation}
Here, $\mathcal{L}^i$ is the loss associated with $i^\mathrm{th}$ cascade (i.e. \emph{companion loss} in DSN terminology) defined by analogy with \eqref{eq:cascadeloss}.
The coefficients $\alpha^i$  regulate the trade-off between losses on different cascades\footnote{One may find $\alpha^i \sim c^i$ to be a relatively good heuristic.}.

In the present work, we find that the objectives defined by \eqref{eq:loss1} and \eqref{eq:loss2} are highly correlated.
However, \eqref{eq:loss2} seems to propagate gradients more effectively, resulting in faster training.

\section{Experiments}

\subsection{Dataset}

As of this writing, no labeled dataset of CMOS images containing true muon tracks is available\footnotemark.
Instead, an artificial dataset was constructed with properties similar to those expected from real data, in order to assess the CNN trigger concept.

\footnotetext{To obtain real data and fully validate performance of the algorithm, an experimental setup with muon scintillators is scheduled this year.}

\begin{figure}[h]
	\centering
	\begin{subfigure}[b]{0.3\linewidth}
		\includegraphics[width=0.49\linewidth]{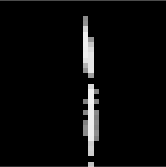}
		\includegraphics[width=0.49\linewidth]{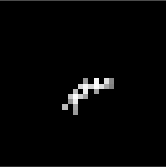}
		\includegraphics[width=0.49\linewidth]{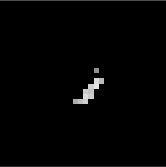}
		\includegraphics[width=0.49\linewidth]{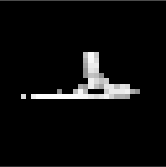}
	     \caption{original traces}
    	 \label{fig:real}
	\end{subfigure}
	\quad
	\begin{subfigure}[b]{0.3\linewidth}
		\centering
		\includegraphics[width=\linewidth]{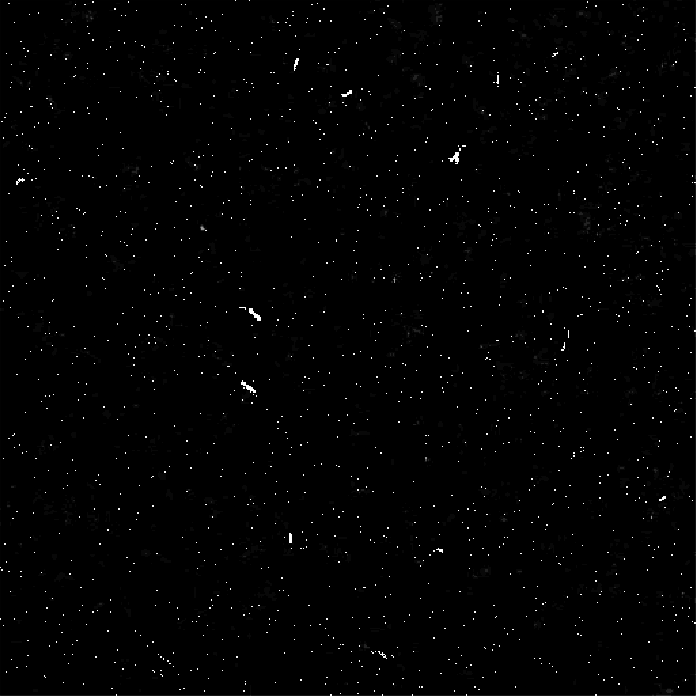}
     	\caption{composition}
    	\label{fig:artificial}
 	\end{subfigure}
    \caption{
    	Test dataset creation steps: \ref{fig:real} selection of bright photon tracks, \ref{fig:artificial}
		track brightness is lowered and superimposed on noisy background.
    }
\end{figure}

To construct the artificial dataset, images were taken from a real mobile phone exposed to radioactive $^{226}$Ra, an X-ray photon source.
These photons interact in the sensor primarily via compton scattering, producing energetic electrons which leave tracks as seen in Fig.~\ref{fig:real}.
These tracks are similar to those expected by muons, the main difference being that the electron tracks tend to be much brighter than the background noise,
rendering the classification problem almost trivial.

Therefore, the selected particle tracks are renormalized such that their average brightness is approximately at the level of the camera's intrinsic noise.
Gloom traces are than superimposed on the background with some Gaussian noise to modeling intrinsic camera sensor noise.
An example of the resulting artificial data is shown in Fig.~\ref{fig:artificial}.
After these measures, the dataset better emulates the case of low-brightness muons, and also forces the classifier to use more sophisticated (geometric) features for classification.

\begin{figure}[h]
	\centering
	\begin{subfigure}[b]{\linewidth}
		\includegraphics[width=0.19\linewidth]{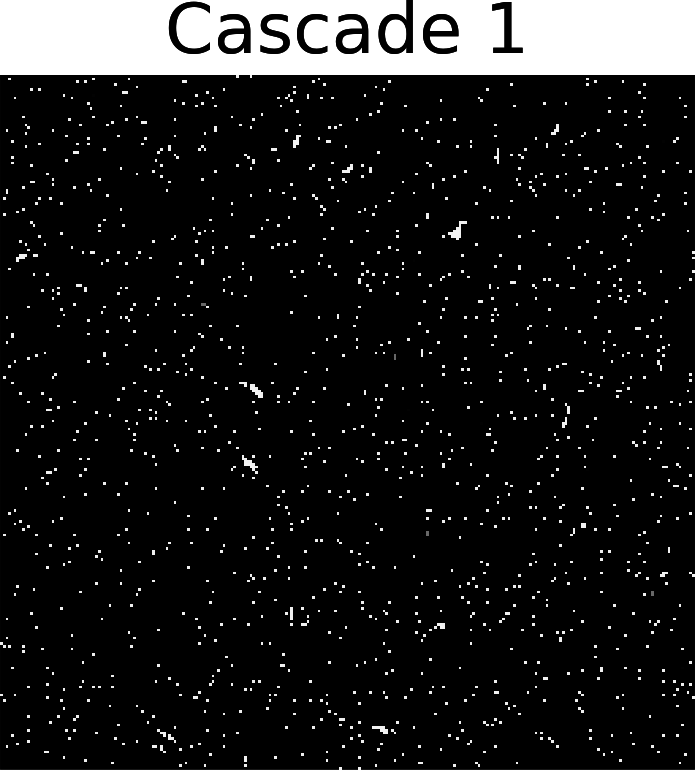}
		\includegraphics[width=0.19\linewidth]{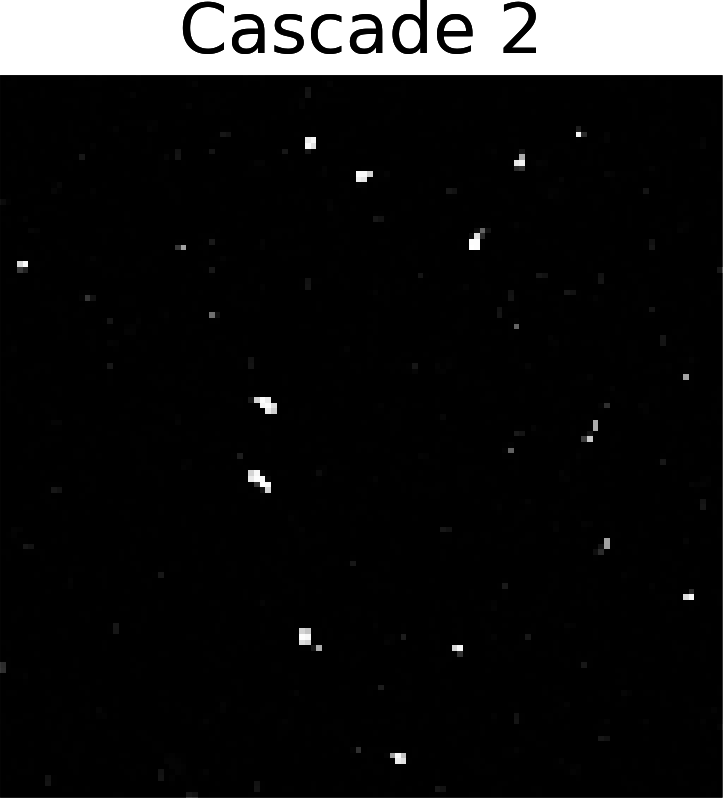}
		\includegraphics[width=0.19\linewidth]{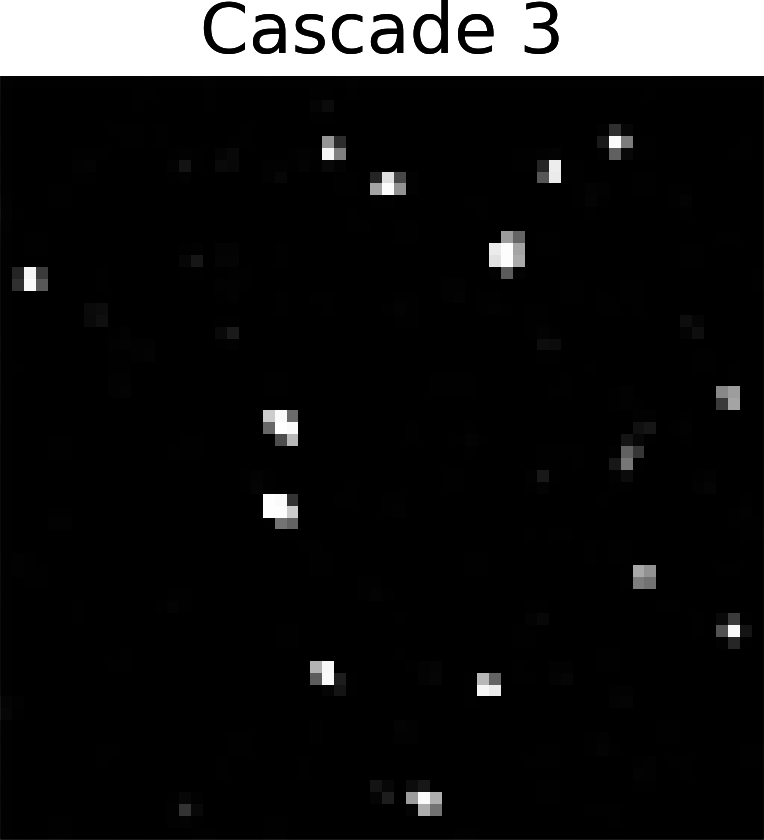}
		\includegraphics[width=0.19\linewidth]{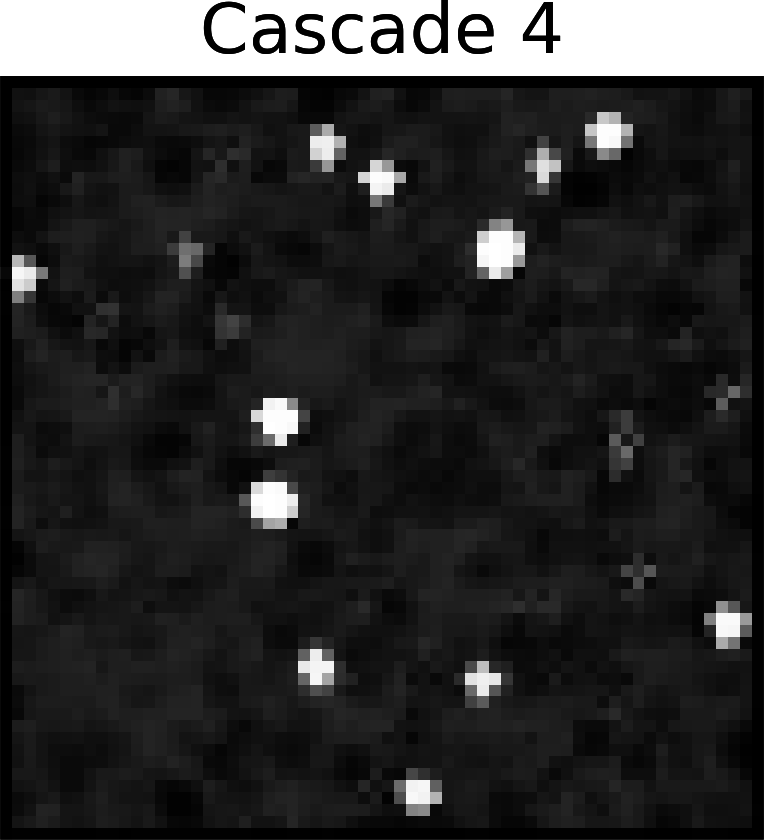}\quad
		\includegraphics[width=0.19\linewidth]{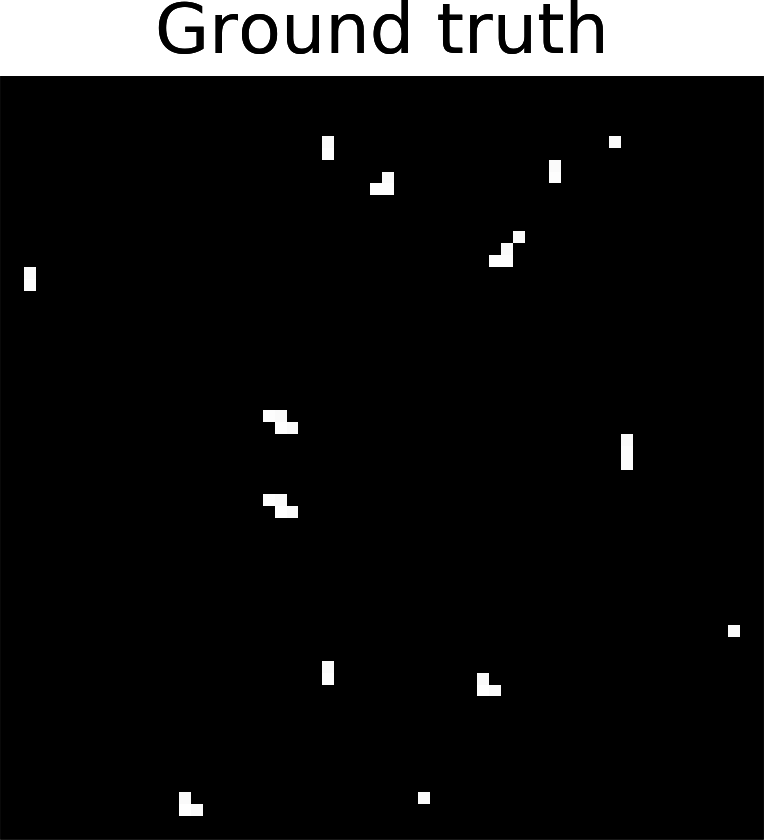}
		\caption{intermediate activation maps and ground truth map}
		\label{fig:iam}
	\end{subfigure}
	\medskip
	\begin{subfigure}[b]{\linewidth}
			\includegraphics[width=0.19\linewidth]{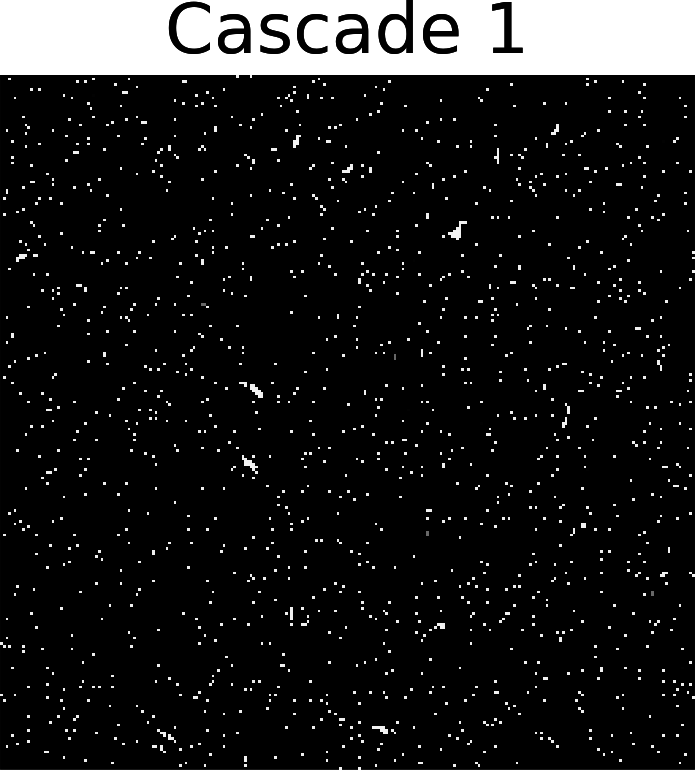}
			\includegraphics[width=0.19\linewidth]{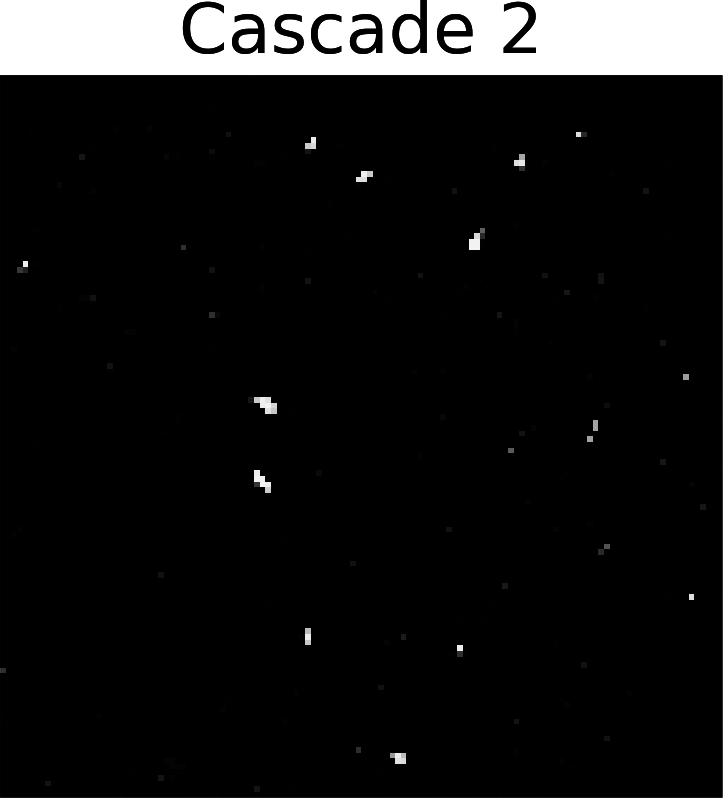}
			\includegraphics[width=0.19\linewidth]{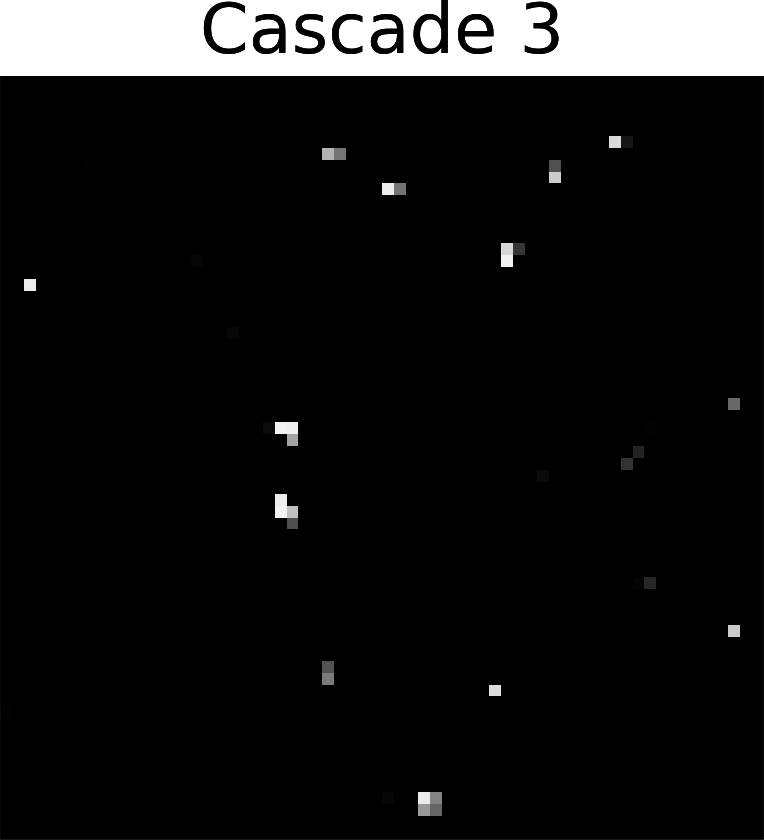}
			\includegraphics[width=0.19\linewidth]{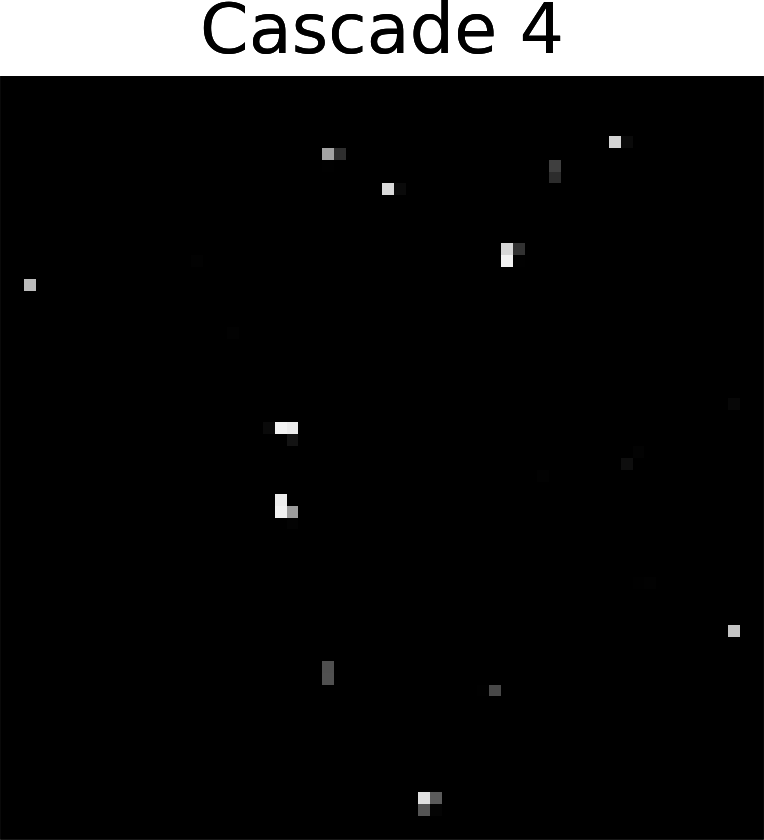}\quad
			\includegraphics[width=0.19\linewidth]{imgs/labels1.png}
		\caption{activation maps and ground truth map}
		\label{fig:am}
	\end{subfigure}
	\medskip
	\begin{subfigure}[b]{\linewidth}
		\includegraphics[width=0.19\linewidth]{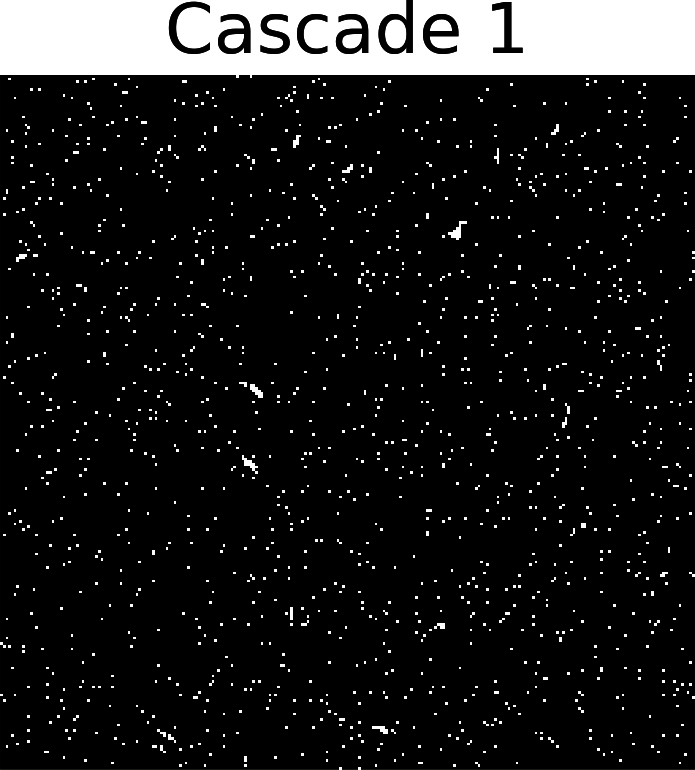}
		\includegraphics[width=0.19\linewidth]{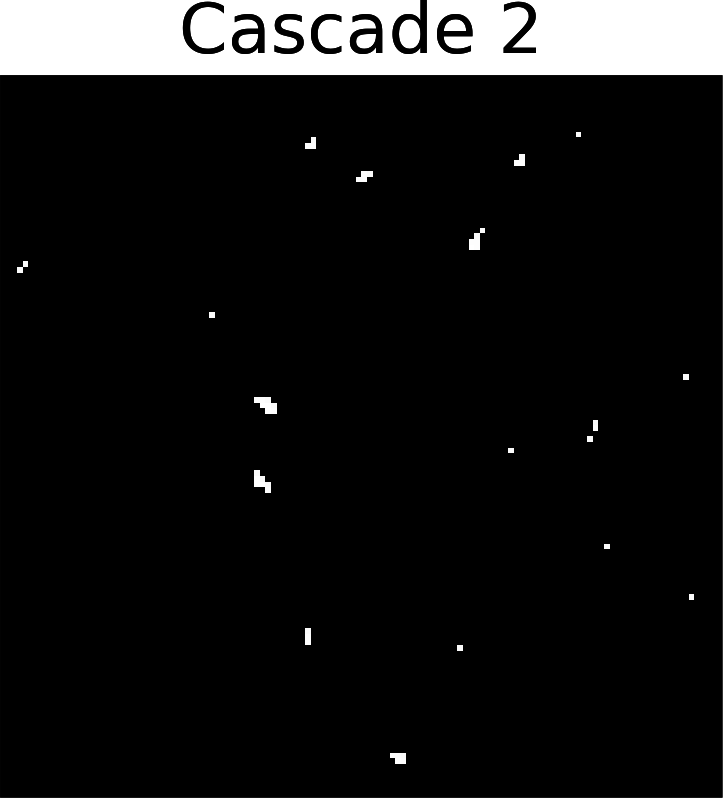}
		\includegraphics[width=0.19\linewidth]{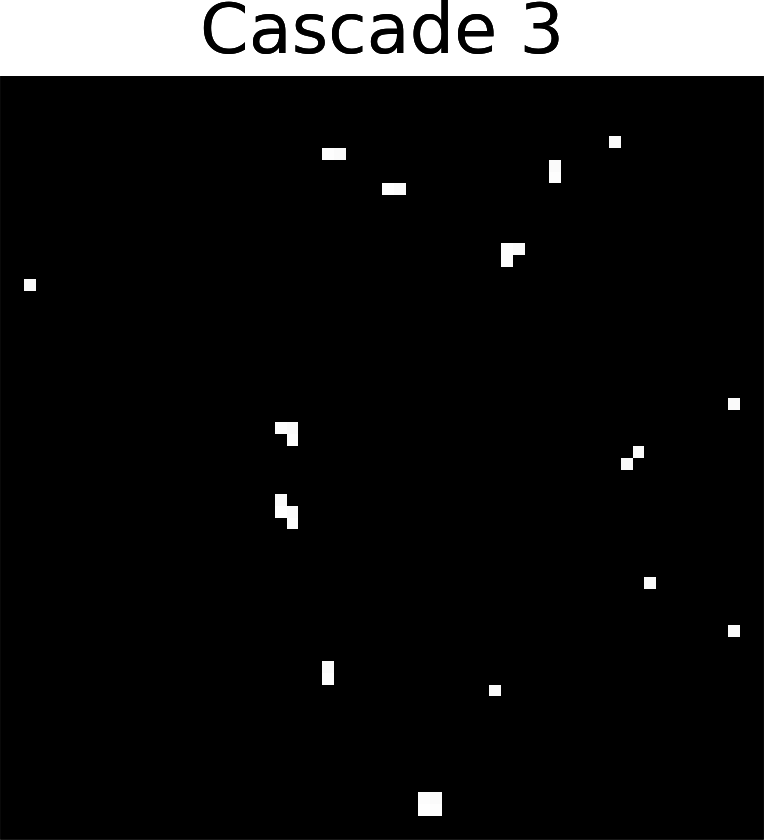}
		\includegraphics[width=0.19\linewidth]{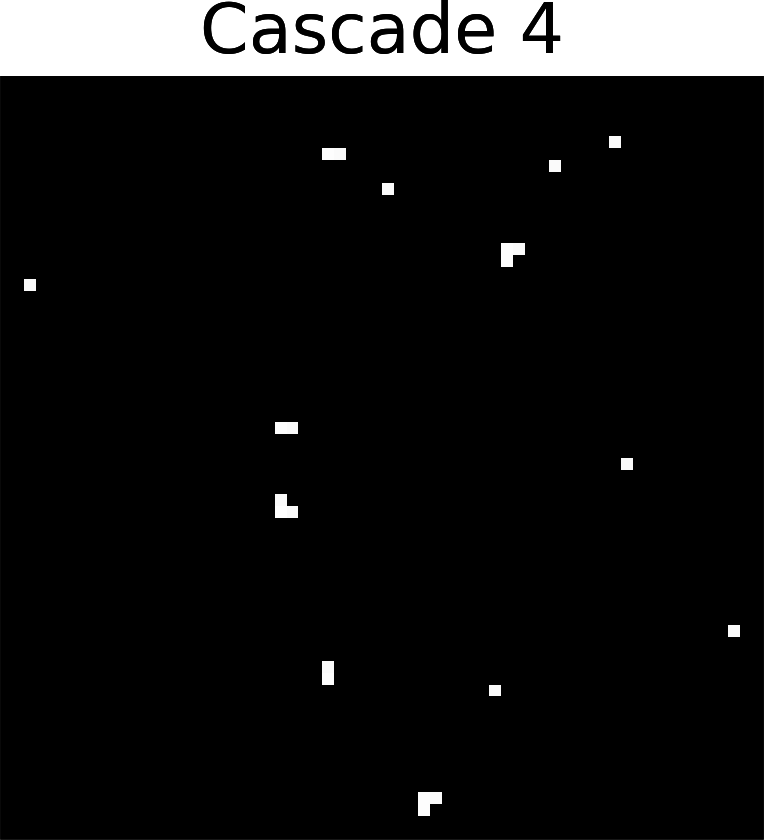}\quad
		\includegraphics[width=0.19\linewidth]{imgs/labels1.png}
		\caption{binary activation maps and ground truth map}
		\label{fig:bam}
	\end{subfigure}
    \caption{
    	Evaluation of the trigger CNN (using the input image from Fig.~\ref{fig:artificial}). Figs.~\ref{fig:iam} and \ref{fig:am} are activation maps for training regime;
    	Fig.~\ref{fig:bam} are binary activation maps for the application regime.
	The resolution of the map is reduced after each cascade to match the downsampling of the internal image representation. 
    }
    \label{fig:actmap}
\end{figure}

\subsection{CNN trigger evaluation}

To evaluate the performance of the method, we consider the case of a CNN trigger with 4 cascades.
The first cascade has a single filter of size $1 \times 1$, equivalent to simple thresholding.
The second, third, and fourth cascades have 1, 3, and 6 filters of size $3 \times 3$, respectively.
Within each cascade, convolutions are followed by $2 \times 2$ max-pooling.

Due to the simple structure of the first cascade, its coefficient $c^1$ from \eqref{eq:complexity} is set to $1$.

As motivated in Sec.~\ref{sec:introduction}, a successful trigger must run rapidly on hardware with limited abilities.
One of the simplest algorithms that satisfies this restriction, thresholding by brightness, is chosen as baseline for comparison.\footnote{In order to obtain comparable results, the output of thresholding was max-pooled to match the size of the CNN trigger output.}
This strategy yields a background rejection rate of around $10^{-2}$ (mainly due to assumptions built in dataset) with perfect signal efficiency.

Two versions of the CNN trigger with average computational costs\footnote{As estimated by \eqref{eq:complexity}.} of $1.4$ and $2.0$ operations per pixel were trained.
This computational cost is controlled by varying coefficients in the loss function \eqref{eq:loss2}.
For each, signal efficiency and background rejection rates at three working points are presented in Table~\ref{tab:res}.
Fig.~\ref{fig:actmap} shows some typical examples of activation maps for different network regimes.

\begin{table}[h]
	\centering
	\begin{tabular}{| p{0.25\linewidth} | c c c | c c c |}
		\hline
		complexity & & $1.4$ op. per pixel & & & $2.0$ op. per pixel & \\
		\hline
		signal efficiency & 0.90 & 0.95 & 0.99& 0.90 & 0.95 & 0.99\\
		\hline
		background rejection & 0.60 & 0.39 & 0.12& 0.65 & 0.44 & 0.15\\
		\hline
	\end{tabular}
	\caption{
		CNN trigger performance for two models with computational costs $1.4$ and $2.0$ operations per pixel.
		Different points for signal efficiency and background rejection were obtained by varying threshold on output of the CNN trigger (i.e. activation map of the last cascade).
	}
	\label{tab:res}
\end{table}

These results indicate a significant improvement of background rejection rate relative to the baseline strategy, even for nearly perfect signal efficiency.

Another performance metric that is interesting to consider is, the normalized computational complexity:
\begin{eqnarray}
	\hat{C} &=& C \cdot \left[\sum^{n}_{i = 1} c^i \sum_{x, y} \left(1 - Y_{x, y}\right)\right]^{-1}\ .\label{eq:normedcomplexity}
\end{eqnarray}

For the models described above, the normalized computational complexity, $\hat{C}$, is around 4-5 percent,
which indicates that a significant amount of computational resources is saved due to lazy application, as compared to a conventional CNN with the same structure.

\section{Conclusion}

We have introduced a novel approach to construct a CNN trigger for fast visual pattern detection of rare events, designed particularly for the use case of fast identification of muon tracks on mobile phone cameras.
Nevertheless, the proposed method does not contain any application-specific assumptions and can be, in principle, applied to a wide range of problems.

The method extends Convolutional Neural Networks by introducing lazy application of convolutional operators, which can achieve comparable performance with lower computational costs.
The CNN trigger was evaluated on an artificial dataset with properties similar to those expected from real data.
Our results show significant improvement of background rejection rate relative to a simple baseline strategy with nearly perfect signal efficiency, while the per-pixel computational cost of the algorithm is increased by less than a factor of $2$.

The effective computational cost is equivalent to 4-5 percent of the cost required by a conventional CNN of the same size.
Therefore the method can enable the evaluation of powerful CNNs in instances where time and resources are limited, or where the network is very large.
This is a promising result for CNNs in many other possible applications, such as very fast triggering with radiation-hard electronics, or power-efficient realtime processing of high resolution sensors.



\section*{References}
\bibliographystyle{iopart-num}
\bibliography{an}
\end{document}